\PassOptionsToPackage{pdfpagelabels=false}{hyperref}
\PassOptionsToPackage{hyphens,spaces,obeyspaces}{url}
\documentclass[shortpaper]{clv2025}
\jvol{}
\jnum{}
\jyear{}

\usepackage{amsmath} 

\usepackage[english]{babel}

\usepackage[utf8]{inputenc}

\usepackage{microtype}

\usepackage{enumitem}

\usepackage{appendix} 

\usepackage{booktabs} 
\usepackage{tabularx}

\usepackage{graphicx}

\usepackage{hyperref}
\usepackage{xcolor}
\definecolor{darkblue}{rgb}{0, 0, 0.5}
\hypersetup{colorlinks=true,citecolor=darkblue, linkcolor=darkblue, urlcolor=darkblue}

\usepackage{multicol} 

\renewcommand{\bibpreamble}{\begin{multicols}{2}}
\renewcommand{\bibpostamble}{\end{multicols}}

\addto\extrasenglish{%
}
\usepackage{etoolbox}  
\makeatletter
\patchcmd{\hyper@makecurrent}{%
    \ifx\Hy@param\Hy@chapterstring
        \let\Hy@param\Hy@chapapp
    \fi
}{%
    \iftoggle{inappendix}{
        \@checkappendixparam{chapter}%
        \@checkappendixparam{section}%
        \@checkappendixparam{subsection}%
        \@checkappendixparam{subsubsection}%
        \@checkappendixparam{paragraph}%
        \@checkappendixparam{subparagraph}%
    }{}%
}{}{\errmessage{failed to patch}}

\newcommand*{\@checkappendixparam}[1]{%
    \def\@checkappendixparamtmp{#1}%
    \ifx\Hy@param\@checkappendixparamtmp
        \let\Hy@param\Hy@appendixstring
    \fi
}
\makeatother

\newtoggle{inappendix}
\togglefalse{inappendix}

\apptocmd{\appendix}{\toggletrue{inappendix}}{}{\errmessage{failed to patch}}
\pretocmd{\appendixsection}{\refstepcounter{section}\addtocounter{section}{-1}}{}{\errmessage{failed to patch}}
\apptocmd{\appendixsection}{\addcontentsline{toc}{section}{#1}}{}{\errmessage{failed to patch}}

\runningtitle{Evaluating Losslessness in Speculative Decoding Under Finite-Precision Inference}
\runningauthor{Koziev}%

\jname{Preprint.}
\jinfo{Evaluating Losslessness in Speculative Decoding Under Finite-Precision Inference}

\begin{document}
\renewcommand{\footmark}{}

\title{Evaluating Losslessness in Speculative Decoding Under Finite-Precision Inference}

\author{Ilya Koziev
, Leonid Sinev
, Ivan Oseledets
}

\affilblock{
}

\maketitle

\begin{abstract}
Lossless speculative decoding is typically defined at the algorithmic level:
a speculative procedure proposes multiple tokens and a verification procedure
is designed to preserve the output trajectory of an autoregressive reference
model exactly. In practical neural inference, however, this guarantee is
implemented using finite-precision floating-point computations, and discrete
token selection can amplify small numerical differences into divergent
generation trajectories. We investigate this distinction using Orthrus, a
hybrid autoregressive-diffusion architecture that performs self-drafting and
self-verification within a frozen autoregressive backbone, as a representative
case study. Across 1{,}190 prompts from 12 domains, exact trajectory matching
under BF16 occurs for only 45\% of the authors' checkpoint generations and
43\% of those from our independently trained model. The probability of
matching is strongly associated with the response-conditional perplexity of
the autoregressive reference, indicating that trajectory divergence is not
uniform across inputs. Despite these divergences, Orthrus does not exhibit
systematic degradation on the evaluated downstream tasks. In contrast, FP32
inference yields exact trajectory matching on all evaluated prompts. These
results demonstrate a gap between algorithmic losslessness and its
implementation under finite-precision arithmetic, and motivate evaluating
lossless speculative decoding at the level of exact generation trajectories
as well as downstream task performance.
\end{abstract}

\section{Introduction}
\label{sec:intro}

Autoregressive (AR) language models have become the dominant paradigm for text generation, but their decoding procedure remains inherently sequential.
Given a generated prefix, the model must compute the next-token distribution before the next token can be appended to the context.
Consequently, generating a sequence of \(N\) tokens generally requires \(N\) decoding steps and repeatedly accesses the growing key-value (KV) cache.
This sequential dependency limits hardware utilization and makes inference increasingly expensive as model and context lengths grow.

Diffusion language models and related parallel decoding methods address this bottleneck by predicting multiple future tokens simultaneously.
However, relaxing the strict autoregressive dependency can introduce discrepancies from the original model distribution.
Orthrus \cite{nguyen-etal-2026-OrthrusMemoryEfficient} proposes an alternative in which, rather than replacing or substantially modifying the pretrained autoregressive model, the method augments a frozen AR backbone with a lightweight diffusion view.
During inference, the AR component constructs the context representation while the diffusion component predicts multiple future tokens in parallel.
An intra-model consensus mechanism then uses the autoregressive view to validate the proposed tokens.
The authors report substantial inference acceleration while claiming that this procedure is strictly lossless, i.e., that the accelerated system preserves the exact predictive behavior of the original autoregressive model.

The present work investigates this losslessness claim experimentally.
We independently implement the Orthrus architecture and training procedure and develop a configurable training framework that enables systematic investigation of training objectives, data distributions, and hyperparameters.
This investigation leads to two observations.

First, we examine the inference efficiency of an independently trained Orthrus model using a teacher-generated distillation corpus.
We construct this corpus by prompting the frozen AR model and recording its greedy-decoded continuations.
The resulting data follows the model's own prediction trajectories.
Our independently trained model achieves a higher TPF point estimate than the released checkpoint in 10 of the 12 evaluation domains.

Second, and more importantly, our experiments reveal that the practical observation of losslessness depends on numerical precision.
The Orthrus consensus mechanism is designed to preserve the autoregressive trajectory at the algorithmic level, but an implementation using finite-precision arithmetic need not reproduce the reference computation exactly.
Because generation is discrete, even small numerical differences can eventually change the selected token and lead to divergent trajectories.

We demonstrate that this effect occurs in practical Orthrus inference.
When generated sequences are compared directly against those of the
corresponding frozen AR model, we observe non-zero rates of output divergence
under BF16 inference.
This finding is particularly relevant because Orthrus explicitly characterizes its generation procedure as strictly lossless.
The observation does not imply that the consensus mechanism is ineffective:
trajectory divergence and downstream task performance are distinct properties.
Rather, it shows that the term ``lossless'' requires a more precise operational definition when applied to neural inference systems implemented with finite-precision arithmetic.

Interestingly, the numerical differences do not necessarily manifest as a
degradation in standard task-level evaluation.
In some experiments, Orthrus obtains benchmark scores that are slightly higher
than those of the corresponding autoregressive model.
This observation further illustrates why benchmark-level equality cannot
establish exact inference equivalence: two systems may obtain similar task
scores while producing different token sequences.

Our contributions are therefore threefold.

First, we provide an independent implementation of Orthrus training and
inference and evaluate it alongside the released checkpoint.

Second, we evaluate an independently trained Orthrus and find a higher TPF point estimate than the released checkpoint in 10 of the 12 evaluated domains.

Third, we show that exact sequence-level equivalence is highly sensitive to
numerical precision for both Orthrus implementations: substantial trajectory divergence occurs under BF16, whereas the same evaluation yields exact trajectory matching under FP32.

Taken together, these results do not undermine the utility of Orthrus as an inference-acceleration technique.
Instead, they clarify an important limitation of strong losslessness claims for neural decoding systems: an algorithm may preserve the intended autoregressive computation in principle while still producing different discrete outputs when implemented with finite-precision arithmetic.
We therefore argue that evaluations of ``lossless'' language-model
acceleration should specify both the operational criterion for equivalence
and the numerical precision under which it is measured.


\section{Related Work}
\label{sec:related_work}

Speculative decoding accelerates autoregressive language-model inference by using a computationally cheaper mechanism to propose multiple future tokens and the target model to verify them in parallel. The framework introduced by \citet{leviathan2023fast} provides an exact sampling procedure that preserves the output distribution of the target model, establishing the basis for subsequent work on lossless speculative decoding.

A line of research removes the need for a separate draft model by using the target model itself for speculation. \citet{zhang2024draft} selectively skips transformer layers during drafting and uses the complete model for verification, while LayerSkip \citep{elhoushi2024layerskip} combines early-exit training with self-speculative decoding. Other approaches augment the target model with lightweight components for parallel prediction. Medusa \citep{cai2024medusa} adds multiple decoding heads and verifies their tree-structured predictions in parallel, while EAGLE \citep{li2024eagle} performs autoregressive drafting at the hidden-feature level. These approaches illustrate several ways of constructing a speculative path while retaining the original autoregressive model as the verifier.

Multi-token prediction (MTP) provides another route to model-internal speculation. \citet{gloeckle2024better} train language models to predict multiple future tokens using auxiliary prediction heads and show that the resulting models can support faster inference through self-speculative decoding. DeepSeek-V3 \citep{liu2024deepseek} adopts a sequential MTP architecture in which additional prediction modules preserve the causal chain across prediction depths and can be used for speculative decoding.

Prior work on lossless speculative decoding commonly defines equivalence in
terms of preserving the target model's output distribution or generated
sequence. Our work focuses on a related but distinct implementation-level question: whether an accelerated model reproduces the exact autoregressive trajectory under finite-precision inference. This question connects to recent studies of numerical reproducibility in LLM inference. \citet{yuan2026understanding} show that changes in numerical precision and other inference-system configurations can alter outputs even under greedy decoding. We investigate this issue specifically in the context of lossless speculative decoding, where exact trajectory equivalence is itself a stated property of the acceleration method.


\section{Experimental Setup}

\subsection{Orthrus Models}

The effects described in \autoref{sec:floating-point-precision} and
\autoref{sec:downstream_task_performance} are observed not only for the original
\texttt{chiennv/Orthrus-Qwen3-1.7B}
checkpoint\footnote{\url{https://huggingface.co/chiennv/Orthrus-Qwen3-1.7B}},
but also for a model of the same capacity that we trained independently using
the procedure described below.
The two models share the same architecture and number of parameters and use the same Orthrus decoding procedure, but differ in their training-data composition, training configuration, and associated inference configuration.
These differences allow us to assess whether the observed effects depend on the specific training setup of the released checkpoint.
We therefore describe our training procedure below, focusing on the aspects relevant to the experiments rather than on implementation details.
    
\subsection{Training Data and Procedure}
\label{sec:training_data}

The training data was constructed by distilling the autoregressive model
\texttt{Qwen/Qwen3-1.7B}~\citep{qwen3technicalreport}.
We collected prompts from a mixture of publicly available instructive datasets hosted on HuggingFace.
The sources and numbers of retained prompts are reported in \autoref{tab:distillation_sources}.
For each prompt, the Qwen3-1.7B model generated a response using greedy decoding.
Each original prompt together with its generated response was then used as a
prompt--response sample for training Orthrus.

\begin{table}[h]
\centering
\small
\begin{tabular}{lr}
\toprule
Dataset & Count \\
\midrule
openbmb/UltraData-SFT-2605                & 1419103 \\
GSAI-ML/ReFusion                                             & 1052176 \\
MBZUAI/LaMini-instruction                                    &  552577 \\
OLMo-Coding/starcoder-python-instruct                        &  284700 \\
meta-math/MetaMathQA                                         &  274711 \\
nvidia/Nemotron-Post-Training-Dataset-v2 &  238290 \\
NTU-NLP-sg/xCodeEval                   &  135841 \\
nvidia/Nemotron-SFT-Instruction-Following-Chat-v2            &  128701 \\
jtatman/python-code-dataset-500k                             &   77618 \\
t-tech/T-Wix                                                 &   69259 \\
ZeroAgency/ru-big-russian-dataset                            &   64841 \\
d0rj/orca-math-word-problems-200k-ru                         &   64621 \\
OpenCoder-LLM/opc-sft-stage2                                 &   48896 \\
sayhan/strix-philosophy-qa                                   &   43837 \\
Post-training-Data-Flywheel/AutoIF-instruct-61k              &   29760 \\
Helsinki-NLP/opus-100                                        &   27614 \\
open-r1/OpenThoughts-114k-math                               &   27298 \\
openbmb/UltraInteract\_sft                                    &   26847 \\
argilla/ifeval-like-data                                     &   17210 \\
camel-ai/math                                                &   13175 \\
allenai/tulu-3-sft-personas-math-grade-filtered              &    8983 \\
bingbangboom/philosophia-QA                                  &    8396 \\
qwedsacf/competition\_math                                    &    7173 \\
Vikhrmodels/GrandMaster-PRO-MAX                              &    7112 \\
MuskumPillerum/General-Knowledge                             &    7057 \\
attn-signs/russian-code                                      &    5491 \\
MexIvanov/CodeExercise-Python-27k-ru                         &    5451 \\
allenai/tulu-3-sft-personas-math-filtered                    &    5315 \\
ajibawa-2023/Python-Code-23k-ShareGPT                        &    4670 \\
teknium/OpenHermes-2.5                                       &    4293 \\
AITISPEC/physics-russian                                     &    2978 \\
RushabhShah122000/python-expert-dataset                      &    2863 \\
MERA-evaluation/MERA                                         &     956 \\
mizinovmv/ru\_ifeval-like-data                                &     629 \\
jondurbin/airoboros-3.2                                      &     582 \\
ise-uiuc/Magicoder-Evol-Instruct-110K                        &     228 \\
attn-signs/russian-easy-instructions                         &     151 \\
greengerong/leetcode                                         &     141 \\
nvidia/Nemotron-SFT-ARC-AGI-v1                               &      69 \\
microsoft/NextCoderDataset                                   &      31 \\
LLiserginov/russian-instructions-10k                         &      25 \\
newfacade/LeetCodeDataset                                    &      21 \\

\bottomrule
\end{tabular}
\caption{The number of prompts taken from each dataset for the training dataset.}
\label{tab:distillation_sources}
\end{table}

Only prompts containing between 50 and 1,000 characters were retained for
distillation.
The resulting dataset contains 4,698,485 prompt--response samples.

\subsection{Training Parameters}
\label{sec:training_params}

Training was performed on eight NVIDIA H100 GPUs using CUDA 13.3.73,
PyTorch 2.13.0, and Transformers 5.8.0.
The training configuration was as follows:
\begin{itemize}
    \item number of epochs: 1;
    \item initial learning rate: $2\times10^{-4}$;
    \item batch size: 10;
    \item loss function: cross-entropy;
    \item block size: 8;
    \item number of blocks: 32;
    \item maximum sequence length: 3,072 tokens.
\end{itemize}

This training configuration differs substantially from that used in the
original Orthrus experiments.

\subsection{Evaluation Dataset}
\label{sec:eval_dataset}

All evaluations of trajectory matching were conducted using a specially curated set of prompts, grouped into 12 text domains with 100 prompts per domain, except for ``gec-en'', which contains 90 prompts.
The domains are described in \autoref{tab:eval_domains}.
This breakdown makes it possible to assess variation in generation trajectories
and their statistical properties across domains.

\begin{table}
    \centering
\caption{Datasets and task domains used for trajectory evaluation.}
\label{tab:eval_domains}
    \begin{tabular}{cllcc}

        \toprule
        \textbf{Domain} & \textbf{Data Source} & \textbf{Task Description} & \textbf{No. prompts} & \textbf{Mean prompt length, toks} \\
        \midrule
    
        code
        & \texttt{me-aas/python-code-dataset-500k}
        & Python code generation in English
        & 100
        & $141.4$ \\

        code-ru
        & \texttt{MERA-evaluation/MERA}
        & Python code generation in Russian
        & 100
        & $179.0$ \\

        creative-en
        & \texttt{IsDeeCee/StoryMaker}
        & English story generation
        & 100
        & $51.5$ \\

        gec-en
        & \texttt{jhu-clsp/jfleg}
        & English grammatical error correction
        & 90
        & $53.1$ \\

        gec-ru
        & \url{https://github.com/ReginaNasyrova/LORuGEC}
        & Human-annotated Russian grammatical error correction
        & 100
        & $68.3$ \\

        math
        & \texttt{HaimingW/math\_train\_decontaminated}
        & English-language mathematical problem solving
        & 100
        & $93.1$ \\

        math-ru
        & \texttt{evilfreelancer/MATH-500-Russian}
        & Russian-language mathematical problem solving
        & 100
        & $91.1$ \\

        poetry-en
        & \texttt{checkai/instruction-poems}
        & English poetry generation
        & 100
        & $62.0$ \\

        poetry-ru
        & n/a (closed sources)
        & Russian poetry generation
        & 100
        & $61.0$ \\

        qa
        & \texttt{smd20/social-engineering-qa-english}
        & General question answering in English
        & 100
        & $33.9$ \\

        qa-ru
        & \texttt{MERA-evaluation/MERA}
        & General question answering in Russian
        & 100
        & $111.4$ \\

        wmt ru-en
        & \texttt{wmt/wmt19}
        & Russian-to-English machine translation
        & 100
        & $105.0$ \\

        \bottomrule
         
    \end{tabular}
    
\end{table}

\subsection{Trajectory Matching}
\label{sec:trajectory_matching_params}

We evaluate losslessness at the level of the generated token trajectory.
For greedy decoding, we define a trajectory as the sequence of token indices
generated from a given prompt until an end-of-sequence token is produced or
the maximum generation length is reached.
A trajectory is considered ``matching'' if it has the same length as the
reference trajectory and all corresponding token indices are identical;
otherwise, it is considered ``diverging''.
Thus, in the greedy setting studied here, exact trajectory matching is our
operational criterion for losslessness.

The results presented in \autoref{sec:floating-point-precision} and
\autoref{sec:ppl} were obtained in the following environment.
We use Python 3.10.12, PyTorch 2.8.0+cu128, CUDA 12.8, and Transformers 5.8.1.
All experiments are performed on an NVIDIA GeForce RTX 3090 GPU with 23~GB of
memory (compute capability 8.6).
Models are evaluated using BF16, FP16, or FP32 precision (\texttt{torch.bfloat16}, \texttt{torch.float16} and \texttt{torch.float32} respectively) with the eager
attention implementation
(\verb|attn_implementation="eager"|).

The decoding algorithm for each Orthrus model uses the block\_size parameter specified in that model's configuration file, corresponding to its training protocol. Thus, inference for the original checkpoint uses a block\_size of 32, while our checkpoint uses a block\_size of 8.

All other model and decoding settings are kept fixed across precision
conditions.

For generation, we use the following arguments of the Transformers
\texttt{generate()} method:
\begin{itemize}
    \item \texttt{max\_new\_tokens=128},
    \item \texttt{do\_sample=False},
    \item \texttt{temperature=0.0}.
\end{itemize}
Thus, all models use greedy decoding without sampling.

\subsection{Downstream Evaluation}
\label{sec:downstream_evaluation_params}

Evaluation on downstream tasks was conducted in the same hardware and software environment described above, using \texttt{lm\_eval==0.4.12}.
All BF16 and FP16 evaluations used identical model, tokenizer, task, and generation configurations and differed only in floating-point precision.
Evaluations were performed using greedy generation with a batch size of 1.
The calculations were performed by calling \texttt{lm\_eval.simple\_evaluate} with the argument \texttt{num\_fewshot=0} for all models.


\section{Results}
\label{sec:results}

\subsection{Inference Efficiency}

We first compare the inference efficiency of the released and independently trained Orthrus models using Tokens Per Forward (TPF), defined as the total number of generated tokens divided by the total number of forward passes during generation.
The results are presented in \autoref{tab:tpf}.
Our model has a higher TPF point estimate in 10 of the 12 domains, while the authors' checkpoint has a higher point estimate in the remaining two.
Because the two models differ in both training data and training configuration, the TPF comparison is descriptive and does not isolate the effect of either factor.

\begin{table}[t]
  \centering
  \caption{Tokens Per Forward across different evaluation domains. Higher values indicate more effective parallel token generation. Values are means with 95\% confidence intervals.}
  \label{tab:tpf}
  \small
  \begin{tabular}{lcc}
    \toprule
    \textbf{Domain} & \texttt{chiennv/Orthrus-Qwen3-1.7B} & \texttt{Orthrus-1.7B-final} \\
    \midrule
    code & $3.24 \pm 0.20$ & $\mathbf{3.44} \pm 0.13$ \\
    code-ru & $3.59 \pm 0.22$ & $\mathbf{3.93} \pm 0.13$ \\
    creative-en & $1.85 \pm 0.09$ & $\mathbf{2.02} \pm 0.10$ \\
    gec-en & $\mathbf{5.00} \pm 0.30$ & $4.23 \pm 0.17$ \\
    gec-ru & $2.48 \pm 0.18$ & $\mathbf{3.23} \pm 0.17$ \\
    math & $\mathbf{8.08} \pm 0.65$ & $4.99 \pm 0.16$ \\
    math-ru & $4.04 \pm 0.28$ & $\mathbf{4.35} \pm 0.14$ \\
    poetry-en & $1.61 \pm 0.04$ & $\mathbf{1.86} \pm 0.04$ \\
    poetry-ru & $1.87 \pm 0.19$ & $\mathbf{2.37} \pm 0.14$ \\
    qa & $2.01 \pm 0.06$ & $\mathbf{2.24} \pm 0.06$ \\
    qa-ru & $1.58 \pm 0.14$ & $\mathbf{2.00} \pm 0.20$ \\
    wmt ru-en & $2.04 \pm 0.09$ & $\mathbf{2.46} \pm 0.12$ \\
    \bottomrule
  \end{tabular}
\end{table}

The independently trained model exhibits similar overall TPF behavior to the released checkpoint while achieving higher TPF in most domains.
We next examine exact trajectory equivalence, which is the operational criterion used here to evaluate losslessness.

\subsection{Trajectory Equivalence Under Different Precisions}
\label{sec:floating-point-precision}

This section analyzes differences between the token trajectories generated by
Orthrus and the original autoregressive model under different floating-point
precision settings, using the inference procedure described in
\autoref{sec:trajectory_matching_params}.

\autoref{tab:trajectory_matching} shows the proportions of Orthrus trajectories
that exactly match the corresponding Qwen3 trajectories and those that contain
at least one divergent token under BF16 inference, aggregated across all
evaluation domains.
The corresponding results broken down by domain are shown in
\autoref{tab:trajectory_matching_domains}.
The results under FP16 inference are presented in
\autoref{tab:trajectory_matching_fp16} and
\autoref{tab:trajectory_matching_domains_fp16}.

\begin{table}[t]
    \centering
    \caption{Orthrus--Qwen3 trajectory matching statistics under BF16 inference. Values are proportions with 95\% Wilson score confidence intervals.}
    \label{tab:trajectory_matching}
    \small
    \begin{tabular}{l c c c}
        \toprule
        \textbf{Model} & \textbf{No. Trajectories} & \textbf{Sequence Match Rate} & \textbf{Diverging Trajectory Rate} \\
        \midrule
        Orthrus-Qwen3-1.7B & 1,190 & $0.45 [0.42, 0.48]$ & $0.55 [0.52, 0.58]$ \\
        Orthrus-1.7B-final & 1,190 & $0.43 [0.40, 0.46]$ & $0.57 [0.54, 0.60]$\\
        \bottomrule
    \end{tabular}
\end{table}

\begin{table}[t]
    \centering
    \caption{Orthrus--Qwen3 trajectory matching statistics under FP16 inference. Values are proportions with 95\% Wilson score confidence intervals.}
    \label{tab:trajectory_matching_fp16}
    \small
    \begin{tabular}{l c c c}
        \toprule
        \textbf{Model} & \textbf{No. Trajectories} & \textbf{Sequence Match Rate} & \textbf{Diverging Trajectory Rate} \\
        \midrule
        Orthrus-Qwen3-1.7B & 1,190 & $0.87 [0.85, 0.89]$ & $0.13 [0.11, 0.15]$ \\
        Orthrus-1.7B-final & 1,190 & $0.87 [0.85, 0.89]$ & $0.13 [0.11, 0.15]$\\
        \bottomrule
    \end{tabular}
\end{table}

\begin{table}[t]
\centering
\caption{Orthrus--Qwen3 trajectory matching rates by domain under BF16
inference. Values are proportions of exactly matching trajectories with 95\% Wilson score confidence intervals.}
\label{tab:trajectory_matching_domains}
\small
\begin{tabular}{lcc}
\toprule
\textbf{Domain} & \textbf{Authors} & \textbf{Ours} \\
\midrule
code        & $0.36 \pm 0.09$ & $0.35 \pm 0.09$ \\
code-ru     & $0.58 \pm 0.09$ & $0.52 \pm 0.10$ \\
creative-en & $0.12 \pm 0.06$ & $0.11 \pm 0.06$ \\
gec-en      & $0.88 \pm 0.07$ & $0.88 \pm 0.07$ \\
gec-ru      & $0.52 \pm 0.10$ & $0.49 \pm 0.10$ \\
math        & $0.54 \pm 0.10$ & $0.54 \pm 0.10$ \\
math-ru     & $0.59 \pm 0.09$ & $0.62 \pm 0.09$ \\
poetry-en   & $0.11 \pm 0.06$ & $0.08 \pm 0.05$ \\
poetry-ru   & $0.14 \pm 0.07$ & $0.11 \pm 0.06$ \\
qa          & $0.25 \pm 0.08$ & $0.17 \pm 0.07$ \\
qa-ru       & $0.73 \pm 0.09$ & $0.70 \pm 0.09$ \\
wmt ru-en   & $0.64 \pm 0.09$ & $0.63 \pm 0.09$ \\
\bottomrule
\end{tabular}
\end{table}

\begin{table}[t]
\centering
\caption{Orthrus--Qwen3 trajectory matching rates by domain under FP16
inference. Values are proportions of exactly matching trajectories with 95\% Wilson score confidence intervals.}
\label{tab:trajectory_matching_domains_fp16}
\small
\begin{tabular}{lcc}
\toprule
\textbf{Domain} & \textbf{Authors} & \textbf{Ours} \\
\midrule
code        & $0.85 \pm 0.07$ & $0.87 \pm 0.07$ \\
code-ru     & $0.92 \pm 0.05$ & $0.94 \pm 0.05$ \\
creative-en & $0.77 \pm 0.08$ & $0.79 \pm 0.08$ \\
gec-en      & $0.99 \pm 0.03$ & $0.99 \pm 0.03$ \\
gec-ru      & $0.94 \pm 0.05$ & $0.91 \pm 0.06$ \\
math        & $0.91 \pm 0.06$ & $0.95 \pm 0.05$ \\
math-ru     & $0.90 \pm 0.06$ & $0.90 \pm 0.06$ \\
poetry-en   & $0.71 \pm 0.09$ & $0.67 \pm 0.09$ \\
poetry-ru   & $0.77 \pm 0.08$ & $0.75 \pm 0.08$ \\
qa          & $0.84 \pm 0.07$ & $0.83 \pm 0.07$ \\
qa-ru       & $0.95 \pm 0.05$ & $0.90 \pm 0.06$ \\
wmt ru-en   & $0.94 \pm 0.05$ & $0.94 \pm 0.05$ \\
\bottomrule
\end{tabular}
\end{table}

The BF16 and FP16 experiments show that neither Orthrus variant consistently reproduces
the autoregressive trajectory exactly.
We next examine whether this discrepancy persists under higher numerical precision.
We therefore repeat the evaluation using FP32 precision for both Orthrus variants and their corresponding reference computations, while keeping the model parameters, decoding procedure, and evaluation prompts unchanged.

Under FP32 inference, Orthrus produces exactly the same token trajectories as
the autoregressive reference on all 1,190 prompts in the trajectory evaluation (\autoref{tab:fp32_trajectory_matching}), whereas the corresponding BF16 configuration exhibits substantial trajectory divergence (\autoref{tab:trajectory_matching}).
The FP32 experiment is restricted to trajectory matching; downstream benchmark evaluation is reported for BF16 and FP16.

\begin{table}[t]
    \centering
    \caption{Orthrus--Qwen3 trajectory matching statistics under FP32 inference. Values are proportions with 95\% Wilson score confidence intervals.}
    \label{tab:fp32_trajectory_matching}
    \small
    \begin{tabularx}{\linewidth}{%
    @{}l%
    >{\centering\arraybackslash}X%
    >{\centering\arraybackslash}X%
    >{\centering\arraybackslash}X@{}%
    }
        \toprule
        \textbf{Model} & \textbf{No. of Trajectories} & \textbf{Sequence Match Rate} & \textbf{Diverging Trajectory Rate} \\
        \midrule
        \texttt{Orthrus-Qwen3-1.7B}
            & 1,190 & $1.00 [0.997, 1.000]$ & $0.00 [0.000, 0.003]$ \\
        \texttt{Orthrus-1.7B-final}
            & 1,190 & $1.00 [0.997, 1.000]$ & $0.00 [0.000, 0.003]$ \\
        \bottomrule
    \end{tabularx}
\end{table}

These observations demonstrate that the practical behavior of Orthrus is
sensitive to numerical precision.
The rate of exact trajectory matching increases when moving from BF16
(\autoref{tab:trajectory_matching}) to FP16
(\autoref{tab:trajectory_matching_fp16}) and reaches 100\% under FP32
(\autoref{tab:fp32_trajectory_matching}).

We therefore attribute the observed trajectory divergence to finite-precision
numerical effects, without attributing it to any particular layer or
computational operation.
Identifying the specific computational stages responsible for these
precision-dependent deviations remains an interesting direction for future
work.

The sensitivity of LLM inference to numerical precision has also been
observed in studies of inference reproducibility, where changes in
floating-point precision and hardware configuration can alter outputs even
under greedy decoding \citep{yuan2026understanding}.

Our setting differs in that we study numerical precision specifically in the
context of lossless speculative decoding: the question is not merely whether
an LLM is reproducible across inference configurations, but whether an
accelerated model reproduces the exact trajectory of its autoregressive
reference.

\subsection{Trajectory Divergence and Reference-Model Perplexity}
\label{sec:ppl}

For a fixed model, prompt, and inference configuration, we next examine whether trajectory matching is systematically associated with characteristics of the generated responses.

A natural hypothesis is that trajectory matching may depend on the reference
model's confidence in the generated response.
We quantify this confidence using the response-conditional perplexity assigned
by the reference Qwen3-1.7B model.
The conditional perplexity values, denoted as PPL (response$\,|$prompt), shown
in the tables are calculated by the Qwen3-1.7B model.
For a prompt $x$ and generated response $y=(y_1,\ldots,y_{|y|})$, the
response-conditional perplexity is defined as
\[
\operatorname{PPL}_{\mathrm{Qwen}}(y \mid x)
=
\exp\left(
-\frac{1}{|y|}
\sum_{t=1}^{|y|}
\log p_{\mathrm{Qwen}}\left(y_t \mid x, y_{<t}\right)
\right).
\]

\begin{table}[t]
    \centering
    \caption{Mean response-conditional PPL for matching and diverging Orthrus trajectories under BF16 inference. Values are means with 95\% Student's $t$-based confidence intervals.}
    \label{tab:mean_ppl}
    \small
    \begin{tabular}{l c c}
        \toprule
        \textbf{Model} & \textbf{Matching trajectories} & \textbf{Diverging trajectories} \\
        \midrule
        Orthrus-Qwen3-1.7B & $1.11 \pm 0.01$ & $1.28 \pm 0.01$ \\
        Orthrus-1.7B-final & $1.10 \pm 0.01$ & $1.28 \pm 0.01$ \\
        \bottomrule
    \end{tabular}
\end{table}

\begin{table}[t]
    \centering
    \caption{Mean response-conditional PPL for matching and diverging Orthrus trajectories under FP16 inference. Values are means with 95\% Student's $t$-based confidence intervals.}
    \label{tab:mean_ppl_fp16}
    \small
    \begin{tabular}{l c c}
        \toprule
        \textbf{Model} & \textbf{Matching trajectories} & \textbf{Diverging trajectories} \\
        \midrule
        Orthrus-Qwen3-1.7B & $1.18 \pm 0.01$ & $1.32 \pm 0.03$ \\
        Orthrus-1.7B-final & $1.18 \pm 0.01$ & $1.32 \pm 0.03$ \\
        \bottomrule
    \end{tabular}
\end{table}

\begin{figure}
    \centering
    \includegraphics[width=0.5\linewidth]{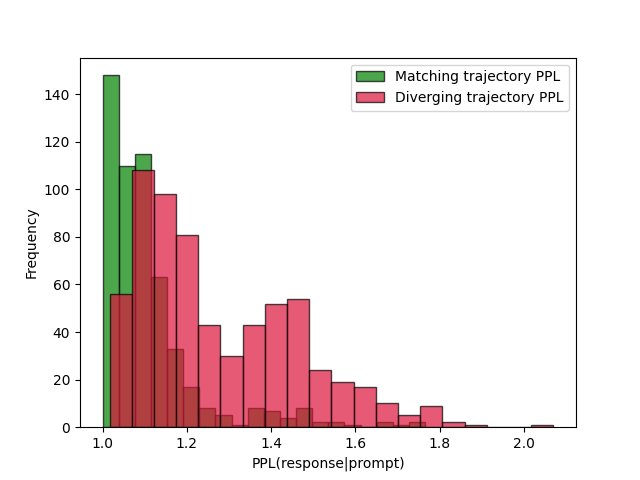}
    \caption{Response-conditional perplexities for matching and diverging trajectories generated by Orthrus-Qwen3-1.7B under BF16 inference.}
    \label{fig:ppl_bf16_authors}
\end{figure}

\begin{figure}
    \centering
    \includegraphics[width=0.5\linewidth]{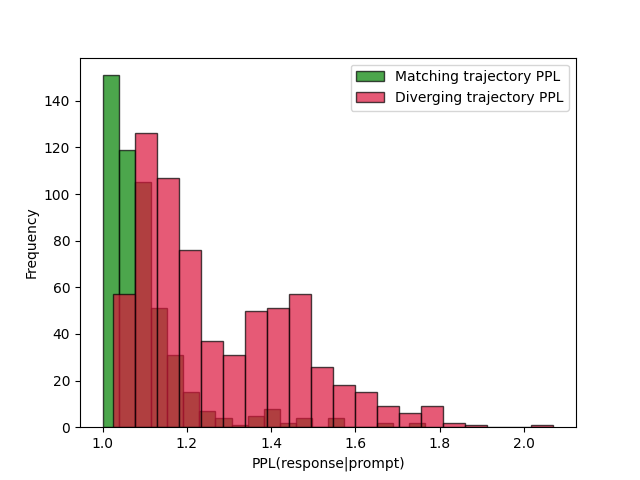}
    \caption{Response-conditional perplexities for matching and diverging trajectories generated by Orthrus-1.7B-final under BF16 inference.}
    \label{fig:ppl_bf16_our}
\end{figure}

\begin{figure}
    \centering
    \includegraphics[width=0.5\linewidth]{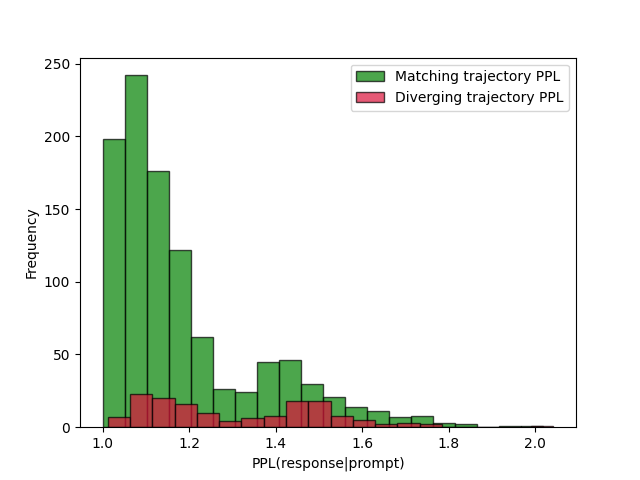}
    \caption{Response-conditional perplexities for matching and diverging trajectories generated by Orthrus-Qwen3-1.7B under FP16 inference.}
    \label{fig:ppl_fp16_authors}
\end{figure}

\begin{figure}
    \centering
    \includegraphics[width=0.5\linewidth]{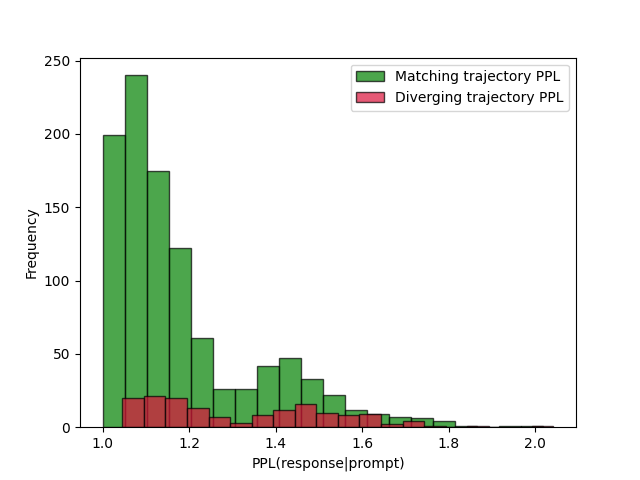}
    \caption{Response-conditional perplexities for matching and diverging trajectories generated by Orthrus-1.7B-final under FP16 inference.}
    \label{fig:ppl_fp16_our}
\end{figure}

Diverging trajectories exhibit higher response-conditional perplexity under
the reference model, as shown in \autoref{tab:mean_ppl} and
\autoref{tab:mean_ppl_fp16} for BF16 and FP16 inference, respectively.
The corresponding PPL distributions are shown in
\autoref{fig:ppl_bf16_authors}, \autoref{fig:ppl_bf16_our},
\autoref{fig:ppl_fp16_authors}, and \autoref{fig:ppl_fp16_our}.
These histograms show differences between the PPL distributions of matching
and diverging trajectories under both precision settings and for both Orthrus
variants.

To test whether this association persists after accounting for response
length and domain, we fit a logistic regression to the trajectory-level
observations, with exact trajectory matching as the binary response and
$\log\operatorname{PPL}(y\mid x)$, response length, and domain as predictors:

\[
\operatorname{logit} P(Y=1)
=
\beta_0+
\beta_1\log\operatorname{PPL}(y\mid x)+
\beta_2L+\gamma_D.
\]
Here, $Y=1$ denotes an exact trajectory match and $\gamma_D$ represents
domain effects.
A negative $\beta_1$ therefore indicates that higher response-conditional
perplexity is associated with a lower probability of exact trajectory
matching.

The model is fitted separately for each Orthrus checkpoint and precision
setting. Logistic regression results for BF16 inference, computed using
\texttt{statsmodels}~\citep{seabold2010statsmodels}, are shown in
\autoref{tab:logistic_regression}.
The results indicate a negative association between response-conditional
perplexity and the probability of exact trajectory matching.
For the
\texttt{chiennv/Orthrus-Qwen3-1.7B} model, the estimated coefficient for
$\log\operatorname{PPL}(y\mid x)$ is $\beta_1=-8.10$ (95\% CI
$[-10.50, -5.71]$, $p=3\times10^{-11}$).
A similar association is observed for \texttt{Orthrus-1.7B-final}, with
$\beta_1=-10.92$ (95\% CI $[-13.60, -8.24]$, $p=1.3\times10^{-15}$).

The logistic regression results for FP16 inference are presented in \autoref{tab:logistic_regression_fp16}.

Thus, for both implementations, responses assigned higher conditional perplexity by the reference Qwen3-1.7B model are associated with a lower probability of exact trajectory matching.
The confidence intervals exclude zero for both Orthrus variants under both precision settings, indicating a negative association after controlling for response length and prompt domain.
These results indicate that trajectory divergence is associated with higher response-conditional perplexity under the reference model rather than being uniformly distributed across inputs.

\begin{table}[t]
    \centering
    \caption{Logistic regression results for the association between response-conditional perplexity and exact trajectory matching for BF16 inference.}
    \label{tab:logistic_regression}
    \small
    \begin{tabular}{lccc}
        \toprule
        \textbf{Model} & $\mathbf{\beta_1}$ & \textbf{95\% CI} & $\mathbf{p}$\textbf{-value} \\
        \midrule
        \texttt{chiennv/Orthrus-Qwen3-1.7B}
            & $-8.10$
            & $[-10.50,\,-5.71]$
            & $3 \times 10^{-11}$ \\
        \texttt{Orthrus-1.7B-final}
            & $-10.92$
            & $[-13.60,\,-8.24]$
            & $1.3 \times 10^{-15}$ \\
        \bottomrule
    \end{tabular}
\end{table}

\begin{table}[t]
    \centering
    \caption{Logistic regression results for the association between response-conditional perplexity and exact trajectory matching for FP16 inference.}
    \label{tab:logistic_regression_fp16}
    \small
    \begin{tabular}{lccc}
        \toprule
        \textbf{Model} & $\mathbf{\beta_1}$ & \textbf{95\% CI} & $\mathbf{p}$\textbf{-value} \\
        \midrule
        \texttt{chiennv/Orthrus-Qwen3-1.7B}
            & $-3.93$
            & $[-6.81,\,-1.05]$
            & $0.007$ \\
        \texttt{Orthrus-1.7B-final}
            & $-3.78$
            & $[-6.60,\,-0.95]$
            & $0.009$ \\
        \bottomrule
    \end{tabular}
\end{table}

Despite the observed trajectory divergence under BF16 and FP16 inference,
these deviations do not translate into systematic degradation in downstream
task performance, as shown in \autoref{sec:downstream_task_performance}.

\subsection{Downstream Task Performance}
\label{sec:downstream_task_performance}

The fact that Orthrus trajectories can diverge from Qwen3 trajectories---even
during greedy generation---does not in itself imply that the resulting responses are defective.
A qualitative inspection of divergent generations revealed cases in which the first differing token leads to a semantically similar continuation.
One such example is shown below.

The prompt was:

``Correct grammar, punctuation, and spelling errors in the text: He said , in other words , that more fluoride may create damage in the human body , specifically the bone .''

Response generated by Qwen3-1.7B:

``He said, in other words, that more fluoride may create damage in the human
body, specifically in the bones.''

Response generated by Orthrus-1.7B:

``He said, in other words, that more fluoride may cause damage in the human
body, specifically in the bones.''

The responses differ by only one word: ``create'' in the Qwen3 response and
``cause'' in the Orthrus response.

To quantitatively assess whether trajectory deviations are associated with
degradation in downstream task performance, we evaluated the models on the
generative tasks GSM8K, HumanEval, and IFEval using the
\texttt{lm-eval-harness} framework~\citep{eval-harness}.
Model and generation parameters are described in
\autoref{sec:downstream_evaluation_params}.

Under BF16 inference, our Orthrus model has higher point estimates than the
autoregressive Qwen3 baseline on all evaluated metrics, as shown in
\autoref{tab:benchmark_comparison}. Under FP16 inference, both Orthrus
variants have lower point estimates than Qwen3 on the evaluated metrics, as
shown in \autoref{tab:benchmark_comparison_fp16}. These differences should
not be interpreted as statistically significant improvements or degradations
given the reported uncertainty.

If the original autoregressive trajectory is taken as the reference behavior,
trajectory deviations might be expected to result in degraded downstream
performance. However, the observed deviations do not consistently correspond
to lower task scores: under BF16, they coincide with higher point estimates
for several evaluated metrics, whereas under FP16 the point estimates are
generally lower. Because the BF16 and FP16 evaluations use the same hardware
and software environment and differ only in floating-point precision, these
results demonstrate that downstream benchmark outcomes can depend on the
numerical precision used during inference. More importantly, the results show that the trajectory divergence explored in \autoref{sec:floating-point-precision} does not by itself imply systematic degradation in downstream task performance.

\begin{table}[t]
    \caption{Comparison of Qwen3-1.7B, the authors' Orthrus-Qwen3-1.7B, and our Orthrus-1.7B-final on \texttt{lm-eval-harness} benchmarks under BF16 inference.}
    \label{tab:benchmark_comparison}
    \centering
    \small
    \begin{tabularx}{\linewidth}{@{}l@{\hspace{0.5em}}Xccc@{}}
        \toprule
        \textbf{Task} & \textbf{Metric} & \textbf{Qwen3-1.7B} & \textbf{Orthrus-Qwen3-1.7B} & \textbf{Orthrus-1.7B-final} \\
        \midrule
        GSM8K
            & exact match (flexible)
            & 0.4003 $\pm$ 0.0135
            & \textbf{0.4238} $\pm$ 0.0136
            & 0.4147 $\pm$ 0.0136 \\
        HumanEval
            & pass@1
            & 0.4024 $\pm$ 0.0384
            & 0.3659 $\pm$ 0.0377
            & \textbf{0.4146} $\pm$ 0.0386 \\
        IFEval
            & prompt-level loose accuracy
            & 0.2015 $\pm$ 0.0173
            & 0.2052 $\pm$ 0.0174
            & \textbf{0.2181} $\pm$ 0.0178 \\
        IFEval
            & prompt-level strict accuracy
            & 0.1682 $\pm$ 0.0161
            & 0.1756 $\pm$ 0.0164
            & \textbf{0.1830} $\pm$ 0.0166 \\
        \bottomrule
    \end{tabularx}
\end{table}

\begin{table}[t]
    \caption{Comparison of Qwen3-1.7B, the authors' Orthrus-Qwen3-1.7B, and our Orthrus-1.7B-final on \texttt{lm-eval-harness} benchmarks under FP16 inference.}
    \label{tab:benchmark_comparison_fp16}
    \centering
    \small
    \begin{tabularx}{\linewidth}{@{}l@{\hspace{0.5em}}Xccc@{}}
        \toprule
        \textbf{Task} & \textbf{Metric} & \textbf{Qwen3-1.7B} & \textbf{Orthrus-Qwen3-1.7B} & \textbf{Orthrus-1.7B-final} \\
        \midrule
        GSM8K
            & exact match (flexible)
            & \textbf{0.4261} $\pm$ 0.0136
            & 0.4230 $\pm$ 0.0136
            & 0.4170 $\pm$ 0.0136 \\
        HumanEval
            & pass@1
            & \textbf{0.4085} $\pm$ 0.0385
            & \textbf{0.4085} $\pm$ 0.0385
            & 0.4024 $\pm$ 0.0384 \\
        IFEval
            & prompt-level loose accuracy
            & \textbf{0.2200} $\pm$ 0.0178
            & 0.2052 $\pm$ 0.0174
            & 0.2107 $\pm$ 0.0175 \\
        IFEval
            & prompt-level strict accuracy
            & \textbf{0.1811} $\pm$ 0.0166
            & 0.1682 $\pm$ 0.0161
            & 0.1738 $\pm$ 0.0163 \\
        \bottomrule
    \end{tabularx}
\end{table}


\section{Limitations and Future Work}
\label{sec:limitations}

Our analysis has several limitations. First, although we establish that the
observed trajectory divergence under BF16 and FP16 disappears under FP32, we do not
identify the specific computational stage responsible for this
precision-dependent behavior. In particular, we do not determine whether the
divergence originates from a particular layer, attention computation, or other
operation in the Orthrus inference procedure. Localizing and characterizing
these numerical effects is an important direction for future work.

Second, our experiments focus on the Qwen3-1.7B-based Orthrus architecture and
a fixed set of evaluation domains.
Although both the released and independently trained checkpoints exhibit the
same qualitative precision-dependent behavior, the experiments do not establish
whether the effect persists at other model scales or in other speculative decoding architectures.

Finally, our downstream evaluation does not examine how trajectory divergence
affects Chain-of-Thought (CoT) reasoning or tool-calling behavior. In complex
reasoning tasks, differences in intermediate generations may propagate through
subsequent reasoning steps, while in tool-using scenarios they may change tool
selection or arguments and thereby alter the subsequent execution trajectory.
Qualitatively assessing these effects is an important direction for future
work.


\section{Conclusion}

We investigated the losslessness claim of Orthrus by directly comparing its generated trajectories with those of the corresponding autoregressive model.
Under BF16 inference, exact trajectory matching occurred for 45\% of the released-checkpoint trajectories and 43\% of those generated by our independently trained model.
Trajectory matching was associated with the response-conditional perplexity of the reference model, while the observed divergences did not correspond to systematic degradation on the evaluated downstream benchmarks.
Crucially, repeating the same trajectory evaluation with FP32 yielded exact matching on all 1{,}190 evaluated prompts.
These results show that the empirical observation of losslessness in Orthrus depends on numerical precision.
More generally, evaluations of lossless language-model acceleration should specify both the operational criterion used to define equivalence and the numerical precision under which that criterion is evaluated.

\phantomsection
\addcontentsline{toc}{section}{References}
\hypersetup{urlcolor=black}
\bibliographystyle{compling-showlinks}
\bibliography{references}

\end{document}